\documentclass[10pt,twocolumn,letterpaper]{article}

\usepackage{iccv}
\usepackage{times}
\usepackage{epsfig}
\usepackage{graphicx}
\usepackage{amsmath}
\usepackage{amssymb}

\usepackage{soul}
\usepackage{url}
\usepackage{mathtools}
\usepackage{bm}
\usepackage{esvect}
\usepackage{enumitem}
\usepackage{subfigure}
\usepackage{multirow}
\usepackage{algorithm}
\usepackage{algorithmic}

\newcommand{\cA}{\mathcal{A}}
\newcommand{\ba}{\mathbf{a}}
\newcommand{\bB}{\mathbf{B}}
\newcommand{\bZ}{\mathbf{Z}}

\newcommand{\bC}{\mathbf{C}}

\newcommand{\cD}{\mathcal{D}}
\newcommand{\cP}{\mathcal{P}}

\newcommand{\bE}{\mathbf{E}}

\newcommand{\bF}{\mathbf{F}}

\newcommand{\cS}{\mathcal{S}}
\newcommand{\bI}{\mathbf{I}}

\newcommand{\bU}{\mathbf{U}}
\newcommand{\bu}{\mathbf{u}}

\newcommand{\bP}{\mathbf{P}}

\newcommand{\bQ}{\mathbf{Q}}

\newcommand{\bR}{\mathbf{R}}
\newcommand{\bS}{\mathbf{S}}
\newcommand{\bs}{\mathbf{s}}

\newcommand{\bV}{\mathbf{V}}

\newcommand{\bx}{\mathbf{x}}

\newcommand{\be}{\mathbf{e}}
\newcommand{\bz}{\mathbf{z}}
\newcommand{\cZ}{\mathcal{Z}}

\newcommand{\cN}{\mathcal{N}}
\newcommand{\bK}{\mathbf{K}}

\newcommand{\bbT}{\mathbb{T}}

\newcommand{\cM}{\mathcal{M}}
\newcommand{\cU}{\mathcal{U}}
\newcommand{\cV}{\mathcal{V}}
\newcommand{\su}{\mathsf{u}}
\newcommand{\sv}{\mathsf{v}}
\newcommand{\sbz}{\mathbf{\mathsf{z}}}
\newcommand{\rmcell}{\mathsf{\tau}}

\usepackage[pagebackref=true,breaklinks=true,letterpaper=true,colorlinks,bookmarks=false]{hyperref}

 \iccvfinalcopy 

\def\iccvPaperID{3183} 

\ificcvfinal\fi

\begin{document}

\title{Event-based Star Tracking via Multiresolution Progressive Hough Transforms}

\author{Samya Bagchi \hspace{1em} Tat-Jun Chin \\
School of Computer Science, The University of Adelaide
}

\maketitle

\begin{abstract}
Star trackers are state-of-the-art attitude estimation devices which function by recognising and tracking star patterns. Most commercial star trackers use conventional optical sensors. A recent alternative is to  use event sensors, which could enable more energy efficient and faster star trackers. However, this demands new algorithms that can efficiently cope with high-speed asynchronous data, and are feasible on resource-constrained computing platforms. To this end, we propose an event-based processing approach for star tracking. Our technique operates on the event stream from a star field, by using multiresolution Hough Transforms to time-progressively integrate event data and produce accurate relative rotations. Optimisation via rotation averaging is then used to fuse the relative rotations and jointly refine the absolute orientations. Our technique is designed to be feasible for asynchronous operation on standard hardware. Moreover, compared to state-of-the-art event-based motion estimation schemes, our technique is much more efficient and accurate.
\end{abstract}

\section{Introduction}\label{sec:intro}

On many space missions, it is vital to estimate the attitude of the spacecraft~\cite{markley14}, which is the 3DOF orientation (roll, pitch, yaw) of the body frame of the spacecraft w.r.t.~an inertial frame, such as the celestial reference frame. The importance of attitude estimation derives from the need to control the bearing of the spacecraft or instruments on board, in order to achieve the mission objectives. Different types of sensors are available for calculating spacecraft attitude, such as sun sensors and  magnetometers. It has been established, however, that star trackers are state-of-the-art in spacecraft attitude estimation~\cite{liebe02}, especially to support high precision orientation determination.

\begin{figure}[h]\centering
\includegraphics[height=15em]{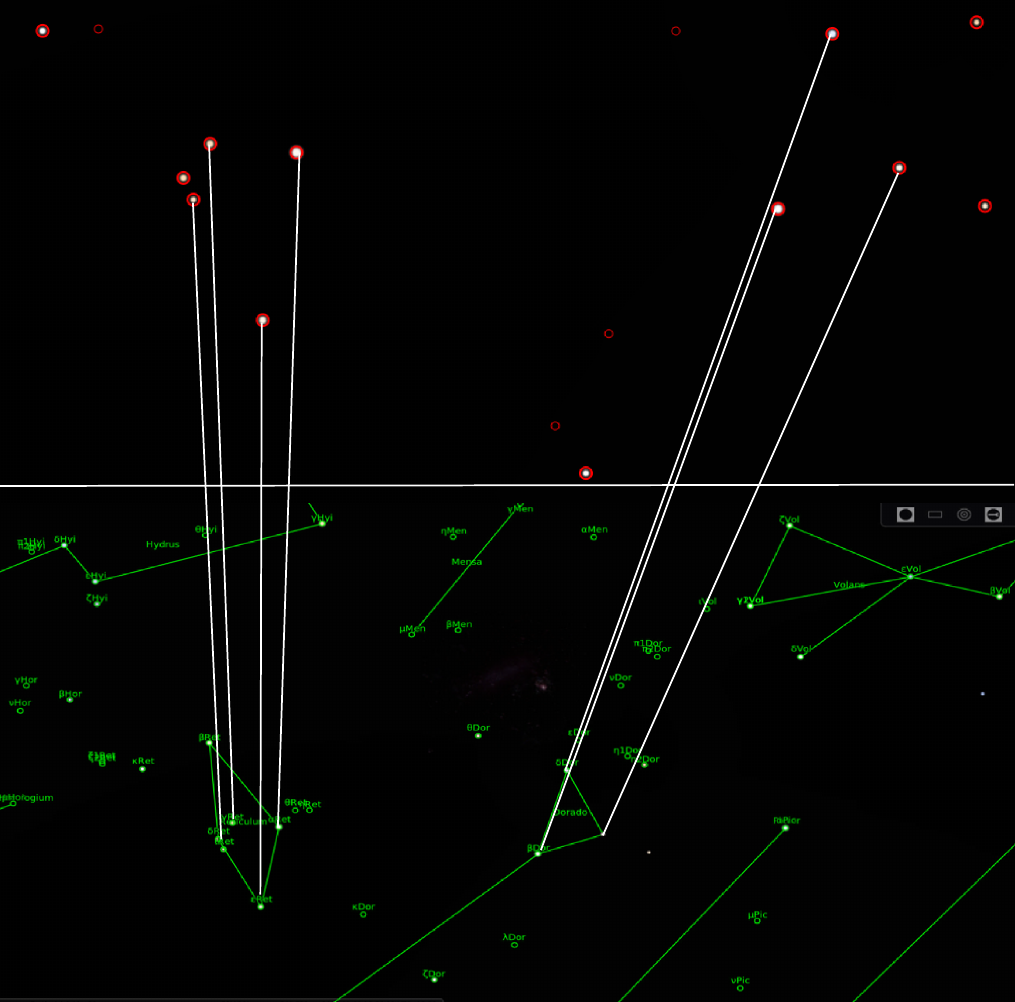}
\caption{In star identification, stars in an input image (top) are identified and matched with known stars in a star catalogue (bottom). This result was obtained using the technique of~\cite{lang10,astrometry}.}
\label{fig:conventional}
\end{figure}

A star tracker is an optical device that estimates spacecraft attitude by recognising and tracking star patterns~\cite[Chap.~4]{markley14}. Let $I$ be an image of a star field captured by the camera of a star tracker. Let $\bF_{ref} \in O(3)$ and $\bF \in O(3)$ respectively define the inertial reference frame, and spacecraft body frame at image $I$. For simplicity, we assume calibrated cameras, thus $\bF$ is also the camera frame at $I$. The attitude at $I$ is defined by the rotation matrix $\bR$, where
\begin{align}
\bF_{ref} = \bR\bF.
\end{align}
In a typical star tracker, the process to determine $\bR$ begins by star identification~\cite{spratling09}: matching the observed stars in $I$ with known stars in a celestial catalogue expressed in $\bF_{ref}$; see Fig.~\ref{fig:conventional}. The matching can be done by comparing local descriptors~\cite{lang10}, geometric voting~\cite{kolomenkin08}, or subgraph matching~\cite{galvizo18}. This establishes a set of 2D-3D correspondences, which are then used to compute $\bR$ via, e.g., SVD~\cite[Chap.~5]{markley14} or more robust techniques~\cite{chin14}. In a practical system, a sequence of attitude estimates over time are then jointly optimised (e.g., using EKF with a kinematic model~\cite[Chap.~6]{markley14}) to yield a set of refined attitudes.

\begin{figure*}[t]\centering
\subfigure[]{\includegraphics[height=11em]{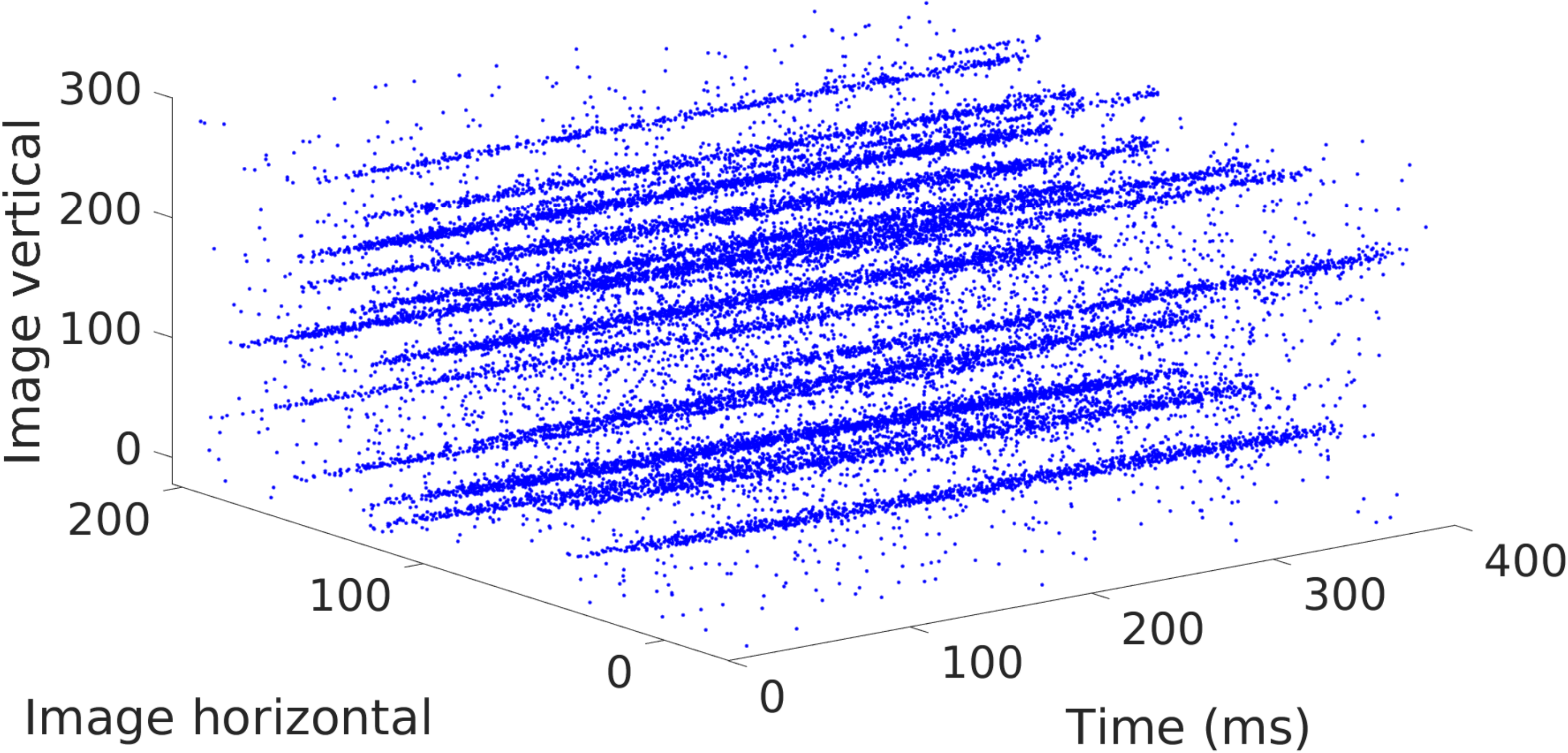}\label{fig:eventdata}}
\subfigure[]{\includegraphics[height=10em]{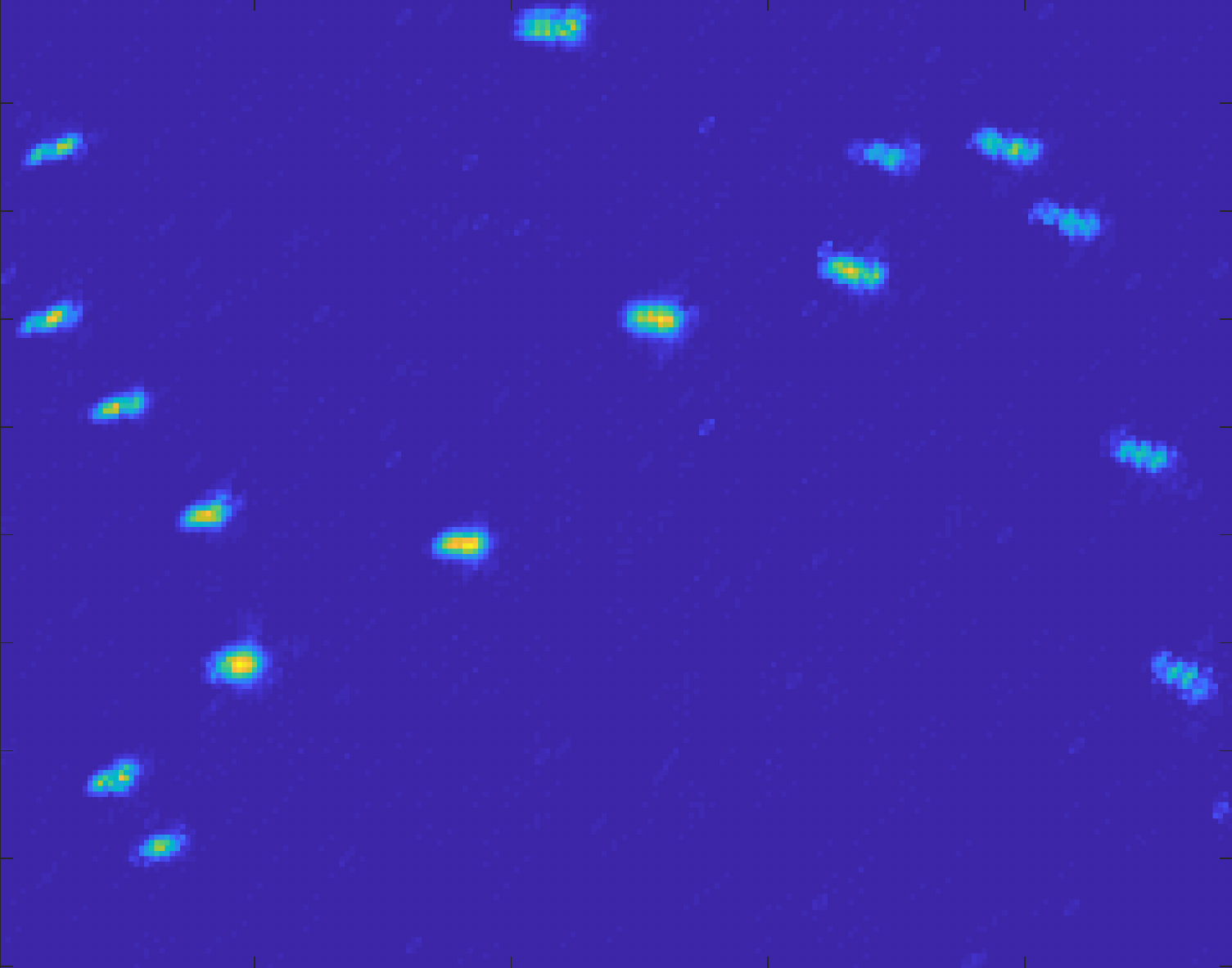}\label{fig:unoptimisedRR}}
\subfigure[]{\includegraphics[height=10em]{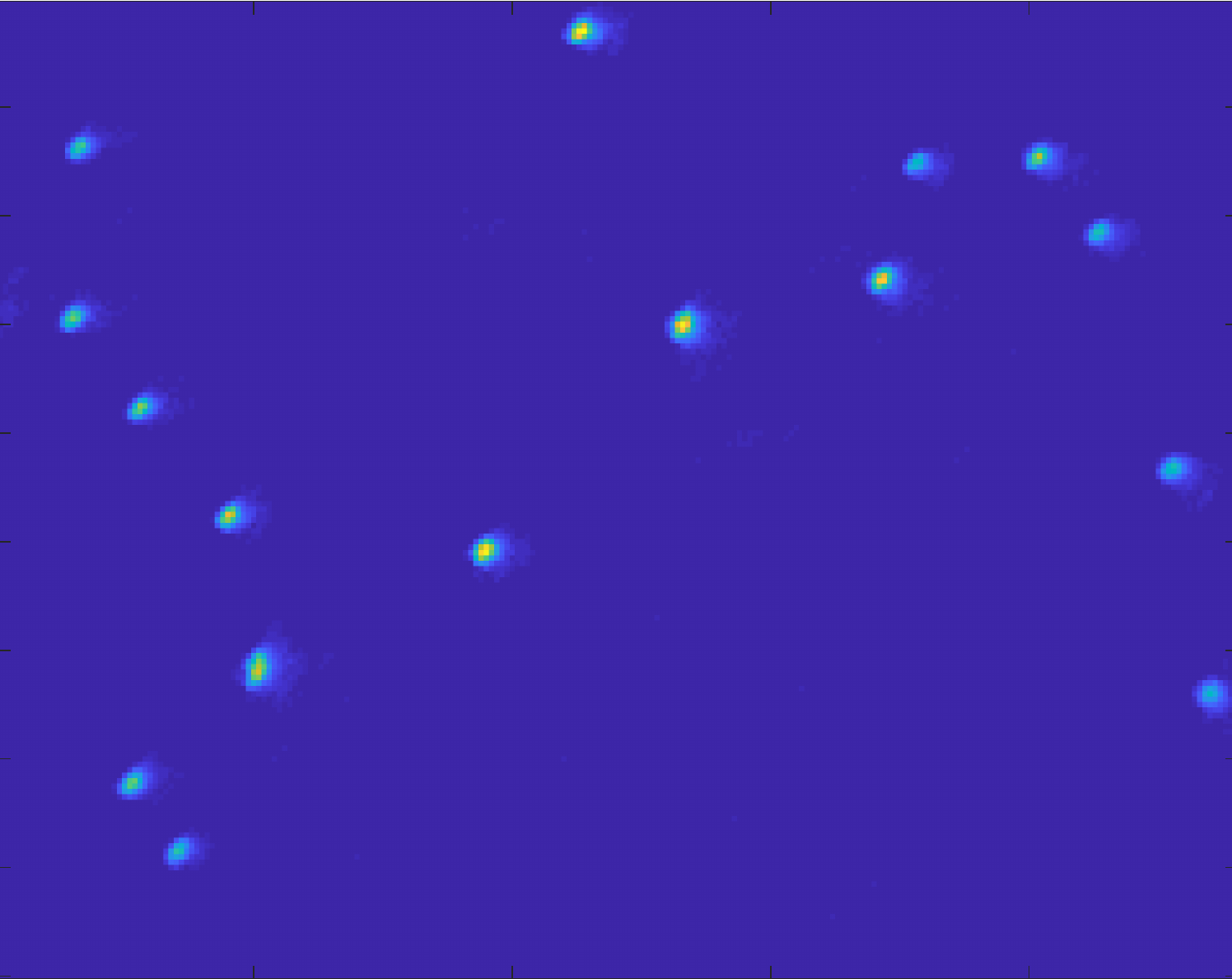}\label{fig:optimisedRR}}
\caption{(a) An event stream from a star field recorded using an event camera under ego-motion (event polarities not plotted; data provided by~\cite{chin18}). (b)(c) Motion compensated event images~\eqref{eq:motcomp} before and after relative rotation estimation by the proposed HT method on the data in (a). Observe that (c) is a sharper star field image than (b), indicating that the proposed method has successfully estimated the ego-motion which generated the event stream. In the proposed relative motion model, the angular velocity in the time period $\bbT$ is assumed constant.}
\end{figure*}

\subsection{Event cameras for star tracking}

Recently, the feasibility of using event cameras for star tracking has been established~\cite{chin18}. Unlike a conventional optical sensor, an event sensor detects intensity changes asynchronously~\cite{lichtsteiner08}. Over a time period $\mathbb{T}$, the output of an event camera is an \emph{event stream} $\cS_{\bbT} = \{\be_i\}^{N}_{i=1}$, where $\be_i = (\bx_i,t_i,p_i)$ is a single event, $\bx_i$ are the 2D pixel coordinates of $\be_i$ on the image plane, $t_i \in \mathbb{T}$ is the time when $\be_i$ occurred, and $p_i \in \{+,- \}$ is the polarity of $\be_i$, i.e., whether it was due to an increase or decrease in intensity. Fig.~\ref{fig:eventdata} illustrates an event stream from observing a star field.

Potential benefits of using event cameras for star tracking are lower power consumption and higher speeds~\cite{chin18}. When observing a star field (a scene with relatively few bright points in front of the space void), a vast majority of the pixel locations will not ``fire", hence, the event sensor may consume less power than a conventional sensor. An event sensor also has a high temporal resolution (e.g., iniVation Davis 240C has $\mu s$ resolution), which could enable higher-speed star tracking for ultra-fine attitude control.

A potential concern is that the sensor becomes ``blind" if the event camera is static w.r.t.~the star field. However, in the context of space missions, the possibility of this is extremely low since motion is almost unavoidable in space. Apart from the orbital motion, a spacecraft (especially a nanosatellite) usually experiences wobbling, which inevitably leads to the generation of event streams.

\vspace{-1em}
\paragraph{Challenges of event-based star tracking}

To reap the potential benefits of event sensors for star tracking, several challenges must be met. First, the fundamentally different kind of data requires new attitude estimation algorithms that do not exist in the literature~\cite{markley14}. Second, the high data rate (e.g., $\mu s$ resolution) demands very efficient algorithms that can process the event stream, and that are also simple enough to be implemented on resource-constrained computing platforms or basic hardware such as FPGA or SoC. 

\subsection{Our contributions}\label{sec:contributions}

We propose a novel event algorithm for star tracking. At the core our technique is the usage of a bank of Hough Transforms (HT)~\cite{matas98} at multiple time resolutions to incrementally process events for calculation of relative attitudes. The relative attitudes serve two purposes: track the camera motion, and integrate event streams for star identification. A rotation averaging formulation is then used to jointly optimise the attitude measurements to yield the final attitude estimates. As we will show in Sec.~\ref{sec:results}, our technique yields much more efficient and accurate star tracking than state-of-the-art event processing methods~\cite{gallego18}.

Crucially, the part of our algorithm (HT for camera tracking) that processes the high-speed event stream is designed to be feasible for parallel computation on an FPGA~\cite{zhou14}. While refinement by rotation averaging is still required in our pipeline, rotation averaging can be solved using cheap algorithms that conduct just a series of small matrix multiplications~\cite[Algorithm 1]{eriksson18}. As we will show in Sec.~\ref{sec:results}, rotation averaging incurs only a small percentage of the total computational effort (i.e., $2s$ over a $45s$ event stream).

\subsection{Previous work}

Event sensing and processing are increasingly popular in computer and robotic vision~\cite{github:event}. Many important capabilities, such as optic flow computation, panomaric stitching, SfM, and SLAM have been shown to be feasible with event cameras. In fact, in settings where high-speed vision is necessary (e.g., UAV flight~\cite{mueggler14}), event cameras have proven to be superior than conventional cameras.

The application of event cameras to space engineering problems is nascent. In~\cite{cohen17}, the feasibility of using event cameras to detect objects in space was established. This was followed by~\cite{cheung18}, where a probabilistic multiple hypothesis tracker (PMHT) was used to track the objects through time. Our work is inspired by~\cite{chin18} who first proposed event cameras for star tracking. However, their algorithm is completely \emph{synchronous}, in that event streams are converted into event images on which frame-based processing is conducted. Moreover, their method depends on solving relatively costly robust point cloud registration, which could be impracticable on resource-constrained platforms. In Sec.~\ref{sec:results}, we will compare the efficacy of our method against~\cite{chin18}.


\section{Event motion model}\label{sec:setting}


Since stars are infinitely far away, the ego-motion that can be resolved from $\mathcal{S}_\bbT$ is restricted to the rotation group, i.e., a continuous rotational motion~\cite{markley14}. For tractability, we first restrict $\bbT$ to be small (e.g., $50-500$ ms) relative to the angular rate of the camera, such that the motion can be modelled as a single rotation $\bR_\bbT$ called the \emph{relative rotation}. Figs.~\ref{fig:unoptimisedRR} and~\ref{fig:optimisedRR} illustrate the feasibility of this model. In Secs.~\ref{sec:formulate} and~\ref{sec:prevevents}, we define $\bR_\bbT$ and survey existing methods to estimate it, before Secs.~\ref{sec:ht} and~\ref{sec:overall} describe our novel technique and overall star tracking architecture.

\subsection{Problem formulation}\label{sec:formulate}

Despite the fundamentally different sensing principle, the pinhole model can be used to describe the imaging process of an event camera~\cite{delbruck2010}. A pixel location $\bx$ of an event $\be$ can thus be backprojected to form a 3D ray
\begin{align}
\vv{\bx} = \frac{\bK^{-1}\bar{\bx}}{\| \bK^{-1}\bar{\bx} \|_2},
\end{align}
where $\bar{\bx} = [\bx^T,1]^T$ is the augmented version of $\bx$, and $\bK$ is the camera intrinsic matrix. Conceptually, the edge point in 3D space that generated $\be$ lies along the ray. Existing calibration techniques for event cameras~\cite[Calibration]{github:event} can be used to obtain $\bK$, thus we assume that $\bK$ is known.


Let $\bs^{(t)}$ be the pixel coordinates of a star at time $t \in \mathbb{T}$. As mentioned above, by restricting $\bbT$ to be small enough, the coordinates of the star at $t = \alpha$ and $\beta$ obey
\begin{align}\label{eq:rotmodel}
\vv{\bs}^{(\alpha)} = \bR_{\bbT} \vv{\bs}^{(\beta)}.
\end{align}
Re-expressing $\bR_{\bbT}$ using axis-angle representation yields
\begin{align}
\vv{\bs}^{(\alpha)} = \exp\left(\theta_{\bbT}\vv{\ba}_{\bbT}\right) \vv{\bs}^{(\beta)}
\end{align}
where $\theta_{\bbT}$ and $\vv{\ba}_{\bbT}$ are respectively the angle and axis of $\bR_{\bbT}$, and $\exp$ denotes the exponential map.

Since the non-spurious events are generated by the edges of star blobs, it is reasonable~\cite{gallego18} to expect that for each non-spurious event $\be_i \in \cS_\bbT$, the following holds
\begin{align}
\vv{\bx}^{(\alpha)}_i = \exp\left(\theta_{\bbT}\vv{\ba}_{\bbT}\right) \vv{\bx}^{(\beta)}_i,
\end{align}
where $\bx^{(\alpha)}_i$ and $\bx^{(\beta)}_i$ are the pixel positions of the edge that generated $\be_i$ if it was observed at $t = \alpha$ and $\beta$. Further, by assuming that $\bbT$ is small enough such that the angular velocity is constant in $\bbT$~\cite{gallego18}, we have that
\begin{align}\label{eq:rotated}
\vv{\bx}^{(\alpha)}_i = \exp\left(\frac{t_i - \alpha}{\beta - \alpha} \theta_{\bbT}\vv{\ba}_{\bbT}\right) \vv{\bx}_i.
\end{align}
Again, Fig.~\ref{fig:optimisedRR} illustrates the feasibility of assuming constant angular velocity when estimating $\bR_{\bbT}$ over small $\bbT$.

\subsection{Previous event processing methods}\label{sec:prevevents}


\paragraph{Using event images}


Following~\cite{chin18}, events near the start and end of $\cS_\bbT$, say $\{ \be_i \in \cS_\bbT \mid \alpha \le t_i \le \alpha + \Delta \}$ and $\{ \be_i \in \cS_\bbT \mid \beta - \Delta \le t_i \le \beta \}$ are used to generate two event images $I_\alpha$ and $I_\beta$~\cite{mueggler15b}. Since the observed stars are points in the images, the rotation $\bR_\bbT$ can be estimated by robustly registering the point clouds~\cite{chetverikov02} in the images.

A disadvantage of~\cite{chin18} is conceptual: converting event streams to images somewhat defeats the purpose of using event sensors. Moreover, robust registration is NP-hard~\cite{chin18b} and can be costly, especially if the number of observed stars is high. In Sec.~\ref{sec:results}, we will compare our method against~\cite{chin18}.

\vspace{-1em}
\paragraph{Contrast maximisation}

The state-of-the-art technique for motion estimation from event streams is contrast maximisation (CM)~\cite{gallego18}, which readily applies to rotation computation. Unlike previous methods (e.g.,~\cite{kim14,cook11,gallego17,reinbacher17}), which rely on panoramic map and optic flow computation, CM is more elegant and shown to be superior.

CM finds the motion parameters that maximise the contrast of the ``motion-compensated" event image
\begin{align}\label{eq:motcomp}
H(\bx \mid \theta_\bbT,\vv{\ba}_\bbT) = \sum_{i=1}^{N} p_i \kappa( \bx - \bx_i^{(\alpha)})
\end{align}
where $\kappa$ is a smoothing kernel (e.g., Gaussian), and
\begin{align}
\bx_i^{(\alpha)} = \frac{\bK^{(1:2)}\vv{\bx}_i^{(\alpha)}}{\bK^{(3)}\vv{\bx}_i^{(\alpha)}}
\end{align}
is the projection of $\vv{\bx}_i^{(\alpha)}$, which is a function of $\theta_\bbT,\vv{\ba}_\bbT$~\eqref{eq:rotated}, and $\bK^{(1:2)}$ and $\bK^{(3)}$ are the first-2 and 3rd rows of $\bK$. The contrast of~\eqref{eq:motcomp} is approximated by the variance $\sigma^2$ of the image, which is a function of $\theta_\bbT,\vv{\ba}_\bbT$:
\begin{align}
\sigma^2(\theta_\bbT,\vv{\ba}_\bbT) = \frac{1}{D} \sum_\bx \left[ H(\bx \mid \theta_\bbT,\vv{\ba}_\bbT) - \mu_H \right]^2,
\end{align}
where $D$ is the number of pixels in the image, and $\bu_H$ is the mean $\frac{1}{D} \sum_{\bx} H(\bx \mid \theta_\bbT,\vv{\ba}_\bbT)$. To estimate $\bR_\bbT$, we solve
\begin{align}\label{eq:contmax}
\max_{\theta_\bbT,\vv{\ba}_\bbT} \sigma^2(\theta_\bbT,\vv{\ba}_\bbT)
\end{align}
under the constraints $0 \le \theta_\bbT\le  \pi$ and $\| \vv{\ba}_\bbT \| = 1$. Problem~\eqref{eq:contmax} is usually solved gradient ascent methods~\cite{gallego18}.

A downside of contrast maximisation is the relatively complex optimisation algorithm required to solve~\eqref{eq:contmax}. In the following, we propose a much more efficient technique for star tracking, and compare against~\cite{gallego18} in Sec.~\ref{sec:results}.

\begin{figure*}[ht]\centering
\subfigure[]{\includegraphics[height=11em]{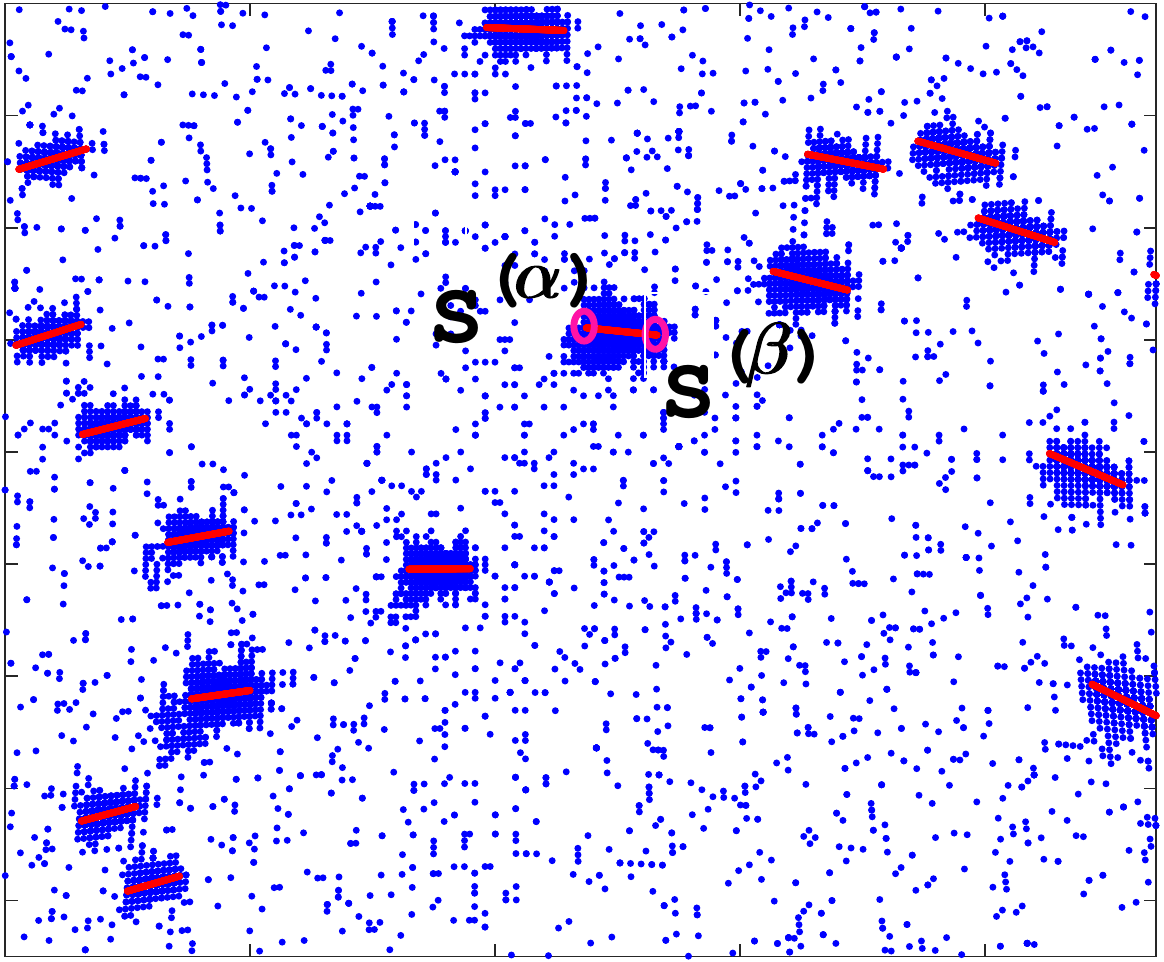}\label{fig:ht2d}}
\hfill
\subfigure[]{\includegraphics[width=0.72\textwidth]{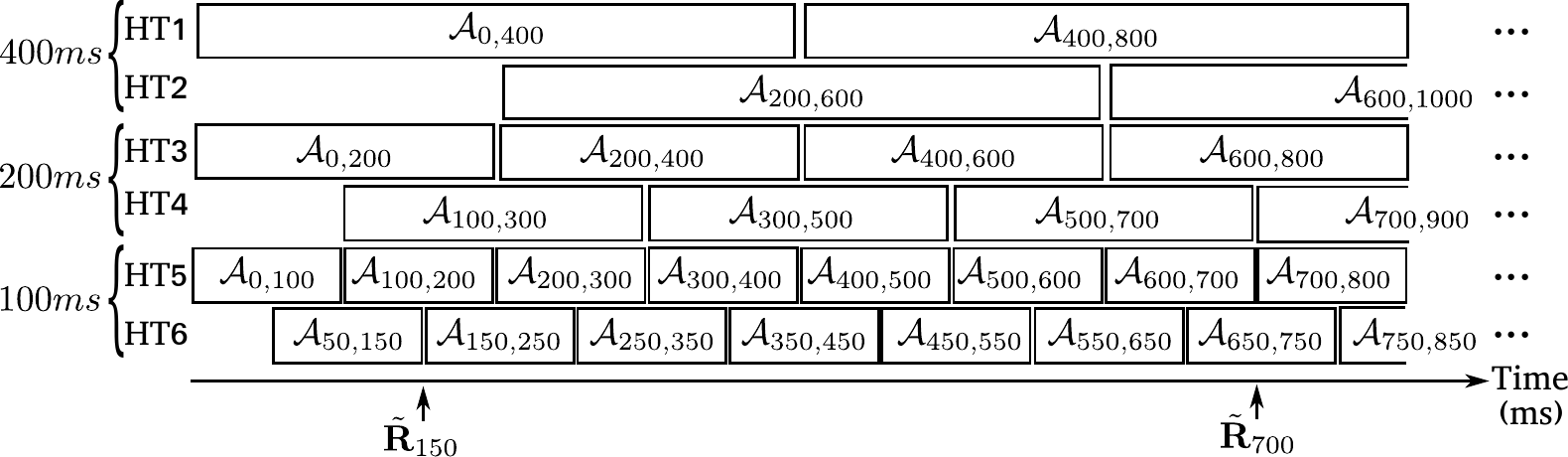}\label{fig:overall}}
\caption{(a) Point correspondences extracted from the lines found in Fig.~\ref{fig:ht2d3d}. (b) Overall architecture of our event-based attitude estimation method. Note that in practice the frequency of absolute attitude estimates $\tilde{\bR}_t$ is much lower than suggested by this diagram.}
\end{figure*}

\begin{figure}[t]\centering
\includegraphics[height=11em]{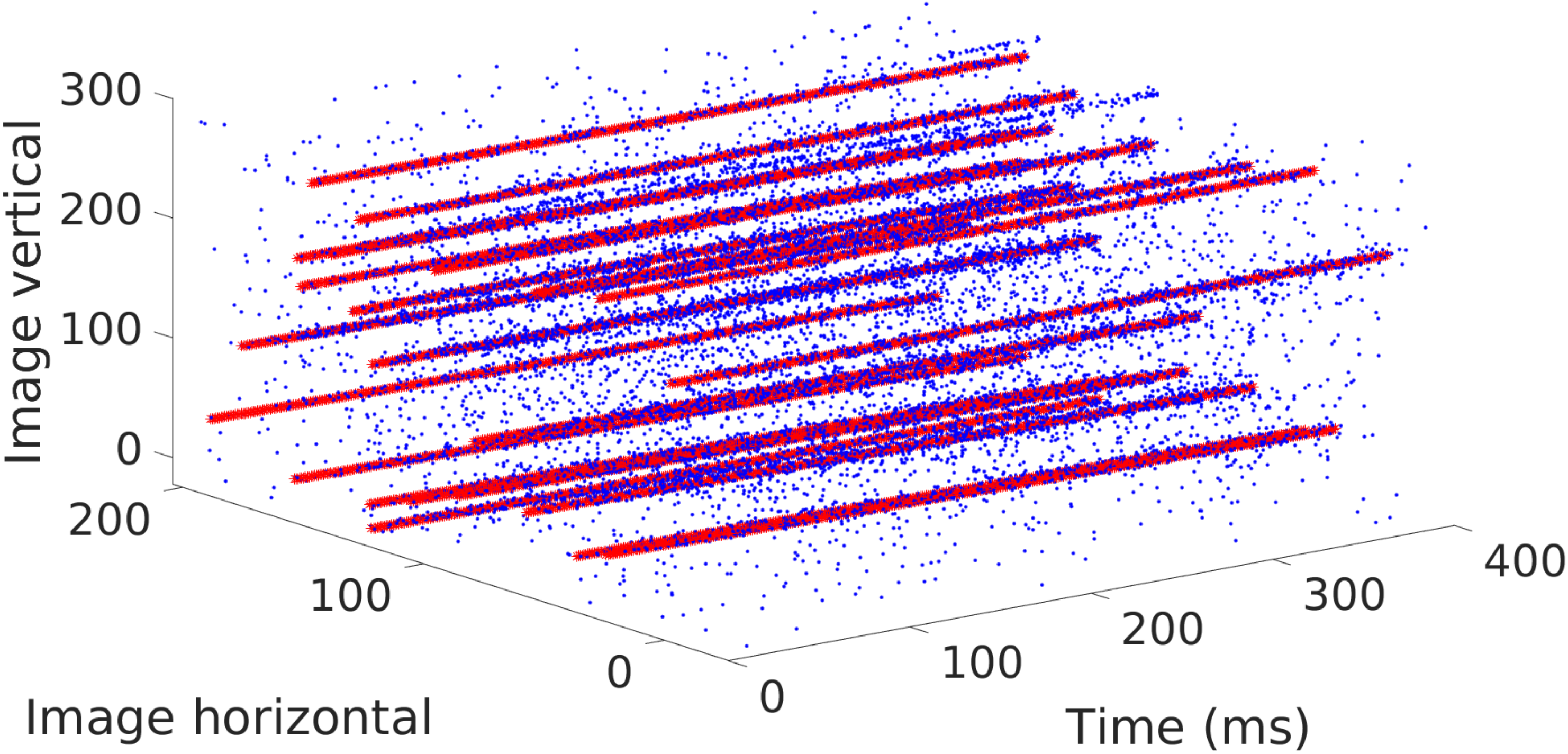}\label{fig:ht3d}
\caption{Lines in event stream found by Algorithm~\ref{alg:ht}.}
\label{fig:ht2d3d}
\end{figure}

\section{Event-based relative rotation estimation}\label{sec:ht}

By ignoring polarity, we convert $\cS_\bbT$ into the 3D point set
\begin{align}
\cP_\bbT = \{ \bz_i \}^{N}_{i=1}, \;\; \text{where} \;\; \bz_i = [\bx_i^T, t_i]^T.
\end{align}
As shown in Fig.~\ref{fig:ht2d3d}, $\cP_\bbT$ consists of points that form linear structures, as well as spurious points (gross outliers). Intuitively, in $\cS_\bbT$ from a short time span, the linear structures consist of events that correspond to stars. Our method for estimating $\bR_\bbT$ from $\cS_\bbT$ exploits this observation. 

Our method does not use the polarity data, though it might be beneficial to do so---we leave this for future work.

\subsection{Relative rotation from 3D lines}

In the 3D spatio-temporal domain $\bz = [\bx^T, t]^T$ obtained by ignoring polarity, a line $\ell$ can be characterised by
\begin{align}\label{eq:line}
\bz_\ell + \lambda \bar{\bz}_\ell \equiv \left[ \begin{matrix} \bx_\ell \\ t_\ell \end{matrix} \right] + \lambda \left[ \begin{matrix} \bar{\bx}_\ell \\ \bar{t}_\ell \end{matrix} \right],
\end{align}
where $\bz_\ell$ is a point on $\ell$, $\bar{\bz}_\ell$ is a unit vector that represents the direction of $\ell$, and $\lambda$ is a scalar. Given noisy points $\cZ \subseteq \cP_{\bbT}$ that belong to $\ell$, by orthogonal least squares~\cite{spath86},
\begin{align}
\bz_\ell = \frac{1}{|\cZ|}\sum_{\bz_i \in \cZ} \bz_i,
\end{align}
i.e., the sample mean of $\cZ$. Let $\bZ$ be the matrix formed by horizontal concatenation of the vectors in $\cZ$, and
\begin{align}\label{eq:meanadj}
\bar{\bZ} = \bZ - \bz_\ell \otimes 1_{1 \times |\cZ|}
\end{align}
be the mean-adjusted matrix. Then, $\bar{\bz}_\ell$ is the most significant left singular vector of $\bar{\bZ}$. In other words, the procedure to estimate $\ell$ is to perform PCA on $\cZ$~\cite{spath86}.

If $\ell$ is due to a star, we seek the ``end points" of $\ell$ at time $\alpha$ and $\beta$. Equating $t_\ell + \lambda t^\prime_\ell$ with $\alpha$ and $\beta$ respectively yields
\begin{align}\label{eq:makematch}
\bs^{(\alpha)} = \bx_\ell + \frac{\alpha - t_\ell}{t^\prime_\ell} \bx^\prime_\ell, \hspace{1em} \bs^{(\beta)} = \bx_\ell + \frac{\beta - t_\ell}{t^\prime_\ell} \bx_\ell^\prime.
\end{align}
These end points are exactly the image coordinates $\bs^{(\alpha)}$ and $\bs^{(\beta)}$ of the star at time $\alpha$ and $\beta$; see~\eqref{eq:rotmodel}. 

If we are able to extract $K$ linear structures from $\cP_\bbT$, by the above calculations we yield the point correspondences
\begin{align}
\{(\bs^{(\alpha)}_k,\bs^{(\beta)}_k)\}^{K}_{k=1}.
\end{align}
Fig.~\ref{fig:ht2d} illustrates correspondences from stars in $\cS_\bbT$. The relative rotation $\bR_\bbT$ can then be estimated from such a correspondence set~\cite{arun87,sorkine17}. Specifically, let
\begin{align}\label{eq:covmat}
\bC = \left[ \vv{\bs}^{(\alpha)}_1, \dots, \vv{\bs}^{(\alpha)}_K \right]\left[ \vv{\bs}^{(\beta)}_1, \dots, \vv{\bs}^{(\beta)}_K \right]^T \in \mathbb{R}^{3\times 3},
\end{align}
and $\bC = \bU\bS\bV^T$ be the SVD of $\bC$. Then,
\begin{align}
\bR_\bbT = \bV\bU^T
\end{align}
is the least squares estimate of the relative rotation.

\subsection{Exact incremental HT for relative rotation}

Our technique to estimate $\bR_\bbT$ finds the linear structures in $\cP_\bbT$ using HT, from which a set of star correspondences are extracted for estimation of $\bR_\bbT$; see Algorithm~\ref{alg:ht}. More details are provided in this section.

Note that although HT has been used previously for event processing~\cite{conradt09,ni12,ni13,seifozzakerini16,valeiras18}, we are the first to conduct event-based relative rotation estimation for star tracking. Second, our algorithm is designed to be feasible for asynchronous operation on standard hardware~\cite{zhou14}. Specifically, each new event $\be_i$ triggers an analytically exact update of the intermediate states, such that when all events are processed, matrix $\bC \in \mathbb{R}^{3\times 3}$ as defined in~\eqref{eq:covmat} is complete. Moreover, each event is processed in constant time (per accumulator cell), which supports high-speed tracking.

\begin{algorithm}[ht]\centering
\begin{algorithmic}[1]
\REQUIRE Time period $\bbT = [\alpha,\beta]$, camera intrinsic matrix $\bK$, Hough cells $\cD \times \cU \times \cV$, peak threshold $\delta$.
\FOR{each Hough cell $\rmcell \in \cD \times \cU \times \cV$}
\STATE $votes^{[\rmcell]} \leftarrow 0,~~~\bz_\ell^{[\rmcell]} \leftarrow \mathbf{0}$,~~~$\bar{\bz}_\ell^{[\rmcell]} \leftarrow \mathbf{0}$.
\ENDFOR
\STATE $\bC \leftarrow \mathbf{0}_{3 \times 3}$.
\WHILE{current time is within $\bbT$}
\FOR{each newly arrived event $\be_i = [\bx_i^T,t_i,p_i]^T$}\label{step:eventstart}
\STATE $\bz_i \leftarrow [\bx_i^T,t_i]^T$.
\FOR{each $\bar{\sbz} \in \cD$}\label{step:loopstart}
\STATE $(u,v) \leftarrow$ Project $\bz_i$ onto plane along $\bar{\sbz}$~\eqref{eq:orthoproj}.
\STATE $[\su,\sv] \leftarrow$ Cell in $\cU \times \cV$ that contains $(u,v)$.
\STATE $\rmcell \leftarrow [\bar{\sbz},\su,\sv]$.
\STATE $votes^{[\rmcell]} \leftarrow votes^{[\rmcell]} + 1$.
\IF{$votes^{[\rmcell]} < \delta$ }
\STATE UpdatePCA$(\bz_\ell^{[\rmcell]}, \bar{\bz}_\ell^{[\rmcell]},\bz_i)$~\eqref{eq:incpca1}--\eqref{eq:incpca2}.
\ELSIF{$votes^{[\rmcell]} = \delta$}
\STATE UpdatePCA$(\bz_\ell^{[\rmcell]}, \bar{\bz}_\ell^{[\rmcell]},\bz_i)$~\eqref{eq:incpca1}--\eqref{eq:incpca2}.
\STATE $\bs^{(\alpha)},\bs^{(\beta)} \leftarrow$ End points of $\bz_\ell^{[\rmcell]} + \lambda\bar{\bz}_\ell^{[\rmcell]}$~\eqref{eq:makematch}.
\STATE $\bC \leftarrow \bC + \vv{\bs}^{(\alpha)}(\vv{\bs}^{(\beta)})^T$.
\ELSE 
\STATE $\bs^{(\alpha)},\bs^{(\beta)} \leftarrow$ End points of $\bz_\ell^{[\rmcell]} + \lambda\bar{\bz}_\ell^{[\rmcell]}$~\eqref{eq:makematch}.
\STATE $\bC \leftarrow \bC - \vv{\bs}^{(\alpha)}(\vv{\bs}^{(\beta)})^T$.
\STATE UpdatePCA$(\bz_\ell^{[\rmcell]}, \bar{\bz}_\ell^{[\rmcell]},\bz_i)$~\eqref{eq:incpca1}--\eqref{eq:incpca2}.
\STATE $\bs^{(\alpha)},\bs^{(\beta)} \leftarrow$ End points of $\bz_\ell^{[\rmcell]} + \lambda\bar{\bz}_\ell^{[\rmcell]}$~\eqref{eq:makematch}.
\STATE $\bC \leftarrow \bC + \vv{\bs}^{(\alpha)}(\vv{\bs}^{(\beta)})^T$.
\ENDIF
\ENDFOR
\ENDFOR\label{step:eventstop}
\ENDWHILE
\STATE $(\bU,\bS,\bV) \leftarrow$ SVD of $\bC \in \mathbb{R}^{3\times 3}$.
\RETURN $\bR_\bbT \leftarrow \bV\bU^T$.
\end{algorithmic}
\caption{Event-based HT to estimate relative rotation.}
\label{alg:ht}
\end{algorithm}


\vspace{-1em}
\paragraph{Hough domain parametrisation}

We follow the Hough parametrisation of~\cite{jeltsch16,dalitz17} for the line $\bz + \lambda\bar{\bz}$ in 3D. The line direction $\bar{\bz} = [\bar{z}_1,\bar{z}_2,\bar{z}_3]^T$, with $\| \bar{\bz} \|_2 = 1$, is discretised as a set $\cD$ of $1281$ vertices of an icosahedron after $4$ recursive steps of polygonal subdivision of each triangular mesh (see~\cite[Fig.~2]{dalitz17}), which is sufficient for our problem.

Instead of discretising $\mathbb{R}^3$ for the point $\bz = [z_1,z_2,z_3]^T$ on the line (which leads to a non-minimal parametrisation), we project $\bz$ along $\bar{\bz}$ onto the plane that passes through the origin that is orthogonal to the line, yielding the 2D point
\begin{align}\label{eq:orthoproj}
\begin{aligned}
u &= \left(1 - \frac{\bar{z}^2_1}{1 + \bar{z}_3} \right)z_1 - \left( \frac{\bar{z}_1\bar{z}_2}{1+\bar{z}_3} \right) z_2 - \bar{z}_1z_3,\\
v &= \left( \frac{\bar{z}_1\bar{z}_2}{1+\bar{z}_3} \right) z_1  + \left(1 - \frac{\bar{z}^2_2}{1 + \bar{z}_3} \right)z_2 - \bar{z}_2z_3;
\end{aligned}
\end{align}
see~\cite{roberts88} for details. The 2D space $(u,v)$ is then discretised as $\cU \times \cV$. By keeping the duration of $\cS_\bbT$ constant (e.g., to $100 m s$) and re-centring $\cP_\bbT$ such that the centroid is at the origin, $\cU \times \cV$ is kept within a fixed bounded region.

\vspace{-1em}
\paragraph{Exact updating}

Each new event $\be_i$ within $\bbT$ votes for the discrete set of line parameters in $\cD \times \cU \times \cV$. Apart from the usual vote accumulator, in each cell $\tau \in \cD \times \cU \times \cV$, we also maintain the \emph{least squares-refined} line parameters $\bz_{\ell}^{[\tau]}$ and $\bar{\bz}_{\ell}^{[\tau]}$ that fit the points that voted for $\tau$.

The key to exact event-triggered updating is to estimate the refined line parameters incrementally. This can be achieved using incremental PCA~\cite{ross08}.  For each $\be_i$ that voted for $\tau$, updating $\bz_{\ell}^{[\tau]}$ is straightforward:
\begin{align}\label{eq:incpca1}
\bz_{\ell}^{[\tau]+} = \frac{votes^{[\tau]}-1}{votes^{[\tau]}} \bz_{\ell}^{[\tau]-} + \frac{1}{votes^{[\tau]}}\bz_i,
\end{align}
where $votes^{[\tau]}$ is the number of votes to $\tau$ \emph{inclusive} of $\be_i$, and $\bz_{\ell}^{[\tau]-}$ and $\bz_{\ell}^{[\tau]+}$ are vector $\bz_\ell^{[\tau]}$ before and after updating.

The trick to update $\bar{\bz}^{[\tau]}$ is to also maintain the left singular vectors $\bP^{[\tau]} \in \mathbb{R}^{3\times 3}$ and non-zero singular values $\mathbf{\Sigma}^{[\tau]} \in \mathbb{R}^{3 \times 3}$ of the mean-adjusted matrix~\eqref{eq:meanadj} of the points that voted for $\tau$ (recall that $\bar{\bz}_\ell^{[\tau]}$ is the left-most vector of $\bP^{[\tau]}$). Given the new $\bz_i$, compute ``difference" matrix
\begin{align}
\hat{\bB} = \left[ \begin{matrix} \mathbf{0}_{3 \times 1} & \sqrt{\frac{votes^{[\tau]}-1}{votes^{[\tau]}}}\left( \bz_i - \bz^{[\tau]-}_\ell\right) \end{matrix} \right] \in \mathbb{R}^{3 \times 2}
\end{align}
and its orthogonal projection onto $\textrm{span}(\bP^{[\tau]})$
\begin{align}\label{eq:qr}
\tilde{\bB} = \textrm{qr}\left( \hat{\bB} - \bP^{[\tau]}(\bP^{[\tau]})^T\hat{\bB} \right) \in \mathbb{R}^{3 \times 2}.
\end{align}
Then, compose the mean-adjusted singular value matrix
\begin{align}
\bE = \left[ \begin{matrix} \mathbf{\Sigma}^{[\tau]} & (\bP^{[\tau]})^T\hat{\bB} \\ \mathbf{0}_{2 \times 3} & \tilde{\bB}^T\left( \hat{\bB} - \bP^{[\tau]}(\bP^{[\tau]})^T\hat{\bB} \right) \end{matrix} \right] \in \mathbb{R}^{5 \times 5}
\end{align}
and compute its SVD
\begin{align}\label{eq:svd2}
\bE = \tilde{\bP}\tilde{\mathbf{\Sigma}}\tilde{\bQ}^T.
\end{align}
Then, $\bP^{[\tau]}$ and $\mathbf{\Sigma}^{[\tau]}$ are updated as
\begin{align}\label{eq:incpca2}
\bP^{[\tau]} = \left[ \begin{matrix} \bP^{[\tau]} & \tilde{\bB} \end{matrix} \right]\tilde{\bP} \;\;\;\; \text{and} \;\;\;\; \mathbf{\Sigma}^{[\tau]} = \tilde{\mathbf{\Sigma}},
\end{align}
and the revised $\bar{\bz}_\ell^{[\tau]}$ is taken as the left-most column of $\bP^{[\tau]}$. Note that the mean-adjusted matrix $\bar{\bZ}$~\eqref{eq:meanadj} (which grows with the number of events) need not be maintined. For more information of the above procedure, see~\cite{ross08}.

In Algorithm~\ref{alg:ht}, if the number of votes in a cell exceeds the pre-determined threshold $\delta$, the least squares-fitted line of the cell is used to extract a star correspondence, following~\eqref{eq:makematch}. However, instead of simply collecting the star correspondences, to facilitate asynchronous operation, the ``covariance" matrix~\eqref{eq:covmat} is directly updated. Note that
\begin{align}
\bC = \sum_{k = 1}^K \vv{\bs}^{(\alpha)}_k(\vv{\bs}^{(\beta)}_k)^T,
\end{align}
hence, the contribution of a new correspondence $(\bs^{(\alpha)},\bs^{(\beta)})$ to $\bC$ can simply be introduced by adding the outer product $\vv{\bs}^{(\alpha)}(\vv{\bs}^{(\beta)})^T$ to $\bC$. If the correspondence from the pre-update line of the cell has contributed to $\bC$, the outer product term due to the old correspondence is first subtracted.

As can be seen in Algorithm~\ref{alg:ht}, as soon as all events within $\bbT$ are processed, matrix $\bC$ is complete and the rotation $\bR_\bbT$ can directly be extracted from $\bC$ via SVD.

\vspace{-1em}
\paragraph{Complexity analysis}

For each incoming event $\be_i$, at most $|\cD|$ Hough cells ($|\cD| = 1281$ in our experiments) are visited (see loop starting in Step~\ref{step:loopstart} in Algorithm~\ref{alg:ht}). As is standard in hardware implementation of HT~\cite{zhou14}, Hough cell voting and its associated operations can be parallelised.

For the per-event computations to be constant time, it remains to show that the least squares line and covariance matrix updating are constant time. The main cost in the former requires the QR decomposition of a $3 \times 2$ matrix $\tilde{\bB}$~\eqref{eq:qr} and SVD of a $5 \times 5$ matrix $\bE$~\eqref{eq:svd2}. Note that although~\eqref{eq:incpca2} seems to increase the sizes of $\bP^{[\tau]}$ and $\mathbf{\Sigma}^{[\tau]}$, only the top-left $3\times 3$ submatrix of the updated results are meaningful for data in 3D space, hence, the sizes of $\bP^{[\tau]}$ and $\mathbf{\Sigma}^{[\tau]}$ can be kept constant at $3 \times 3$. Lastly, updating the $3 \times 3$ covariance matrix $\bC$ by adding/subtracting the outer product of a single operation is clearly constant time. Note that there are also no ``racing" issues with updating $\bC$ in parallel, since each cell contributes to $\bC$ independently.

\section{Overall architecture for attitude estimation}\label{sec:overall}

Fig.~\ref{fig:overall} shows the overall architecture of our event-based attitude estimation technique. Symbol $\cA_{\alpha,\beta}$ indicates the instance of the proposed HT (Algorithm~\ref{alg:ht}) for the time period $[\alpha,\beta]$. We use a bank of HTs at multiple time resolutions (specifically, $400 ms$, $200 ms$, $100 ms$; these can be changed to accommodate data with different angular rates) to track rotational motions over longer time periods.

For each time resolution, two HT instances separated by a stride half of the resolution (e.g., for $400 ms$, the two HTs differ by a stride of $200 ms$) are employed, leading to a total of six HTs. Each HT processes incoming events asynchronously (Sec.~\ref{sec:ht}). When the period of a HT is finished, a relative rotation over that period is returned (e.g., $\cA_{100,300}$ outputs the relative rotation $\bR_{100,300}$), and the HT is immediately restarted for the next period (e.g., $\cA_{300,500}$).

A relative rotation describes the relative attitude change across a time period. In star tracking, however, the quantity of interest is the ``absolute" attitude~\cite{markley14} (see Sec.~\ref{sec:intro}). Moreover, simply chaining the relative attitudes will lead to drift errors. Thus, in our pipeline, motion-compensated event images~\eqref{eq:motcomp} are generated using the HT results (we use the $100 ms$ HTs for this purpose) and subject to star identification and absolute attitude estimation using ~\cite{lang10,astrometry}. The frequency of absolute attitude estimation is much lower, due to the higher cost of star identification, e.g., minutes.


The small set of absolute attitudes are used to ``ground" the relative rotations via a rotation averaging process, which also denoises the measured quantities and produce the final set of refined (absolute) attitude estimates. Let
\begin{align}
\{ \tilde{\bR}_{\alpha,\beta} \}_{\langle \alpha, \beta \rangle \in \cN}
\end{align}
be the set of measured relative attitudes, where $\cN$ encodes the periods $\bbT = [\alpha,\beta]$ processed by a HT instance, and let
\begin{align}\label{eq:grounding}
\{ \tilde{\bR}_{\gamma} \}_{\gamma \in \cM}
\end{align}
be the set of absolute attitude measurements. Note that $\cN$ must be a connected graph, in that it is always possible to find a path between any two ``time stamps" $\alpha$ and $\beta$. Define
\begin{align}
\mathcal{T} = \{ 0, \Delta t, 2\Delta t, 3\Delta t, \dots \}
\end{align}
be the time stamps of the set of absolute attitudes that will be optimised by rotation averaging, where the value of $\Delta t$ is a common denominator of the time resolutions of the HTs employed. For the architecture in Fig.~\ref{fig:overall}, we set $\Delta t = 50 ms$, which also ensures $\cM \subset \mathcal{T}$.

\begin{figure*}[t]\centering
\subfigure[Sequence 1]{\includegraphics[width=0.31\textwidth, height=0.1775\textheight]{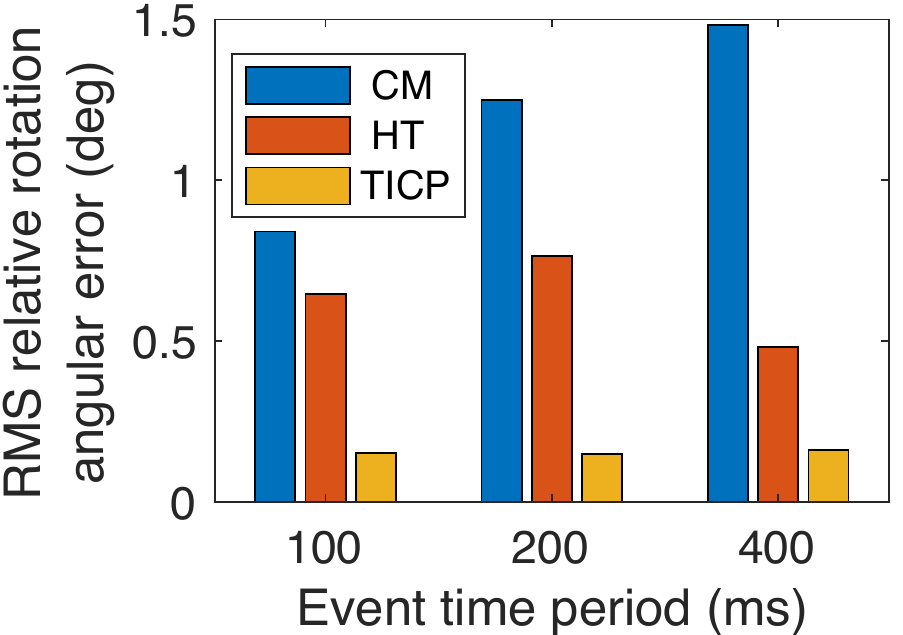}\label{fig:adelaideRR_}}\hspace{1em}
\subfigure[Sequence 2]{\includegraphics[width=0.31\textwidth, height=0.1775\textheight]{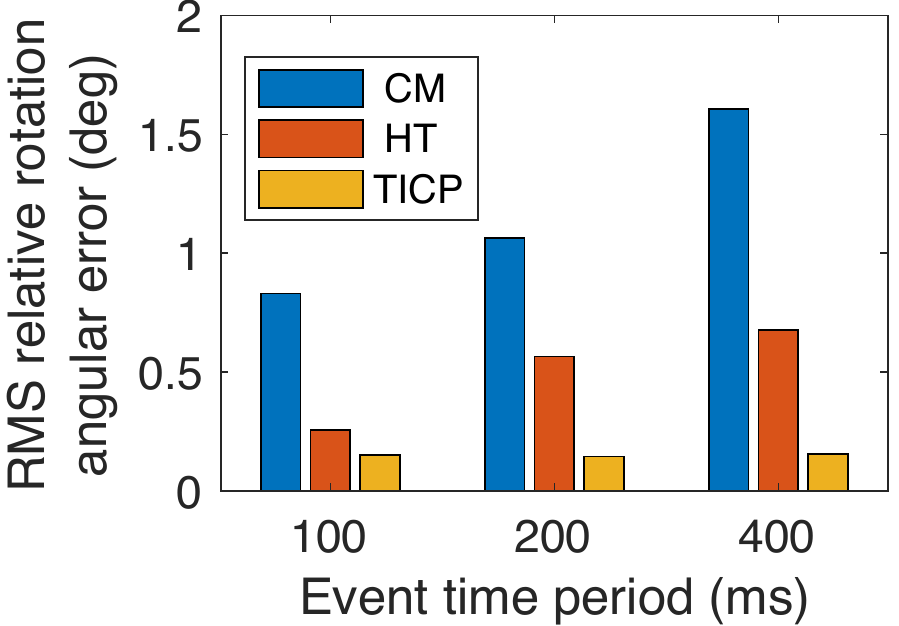}\label{fig:capetownRR_}}\hspace{1em}
\subfigure[Sequence 3]{\includegraphics[width=0.31\textwidth, height=0.1775\textheight]{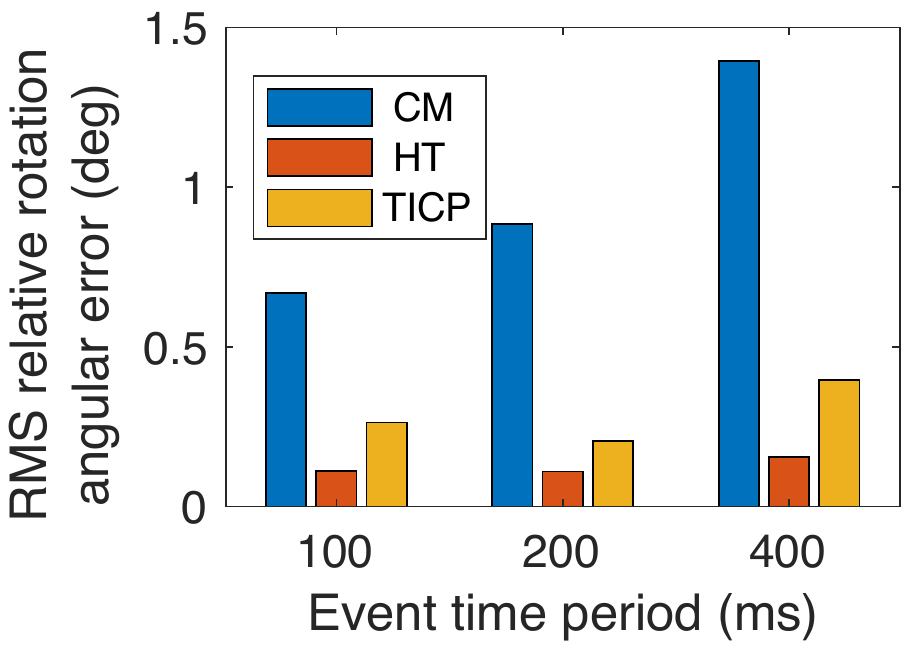}\label{fig:ecuadorRR_}}
\subfigure[Sequence 4]{\includegraphics[width=0.31\textwidth, height=0.1775\textheight]{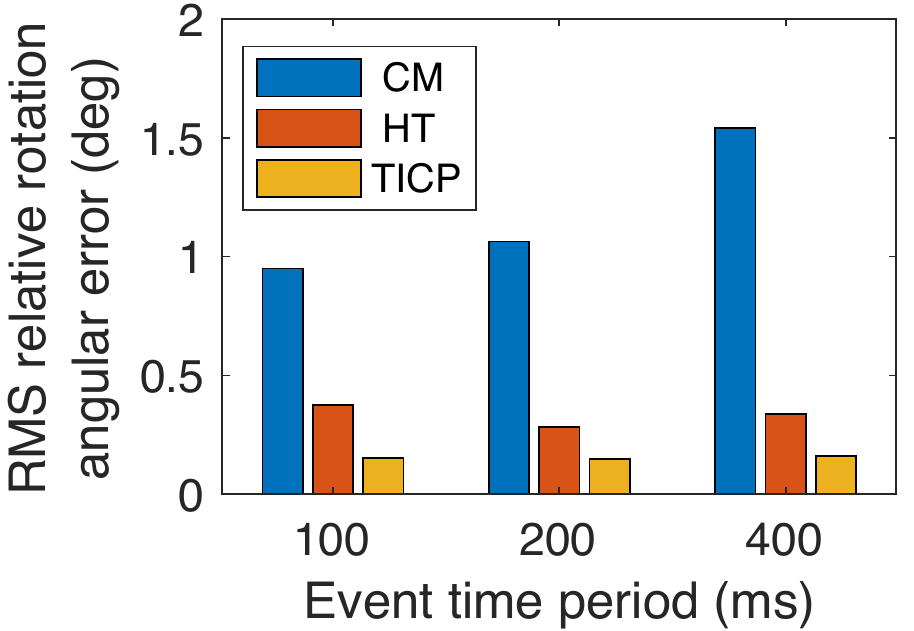}\label{fig:aucklandRR_}}\hspace{1em}
\subfigure[Sequence 5]{\includegraphics[width=0.31\textwidth, height=0.1775\textheight]{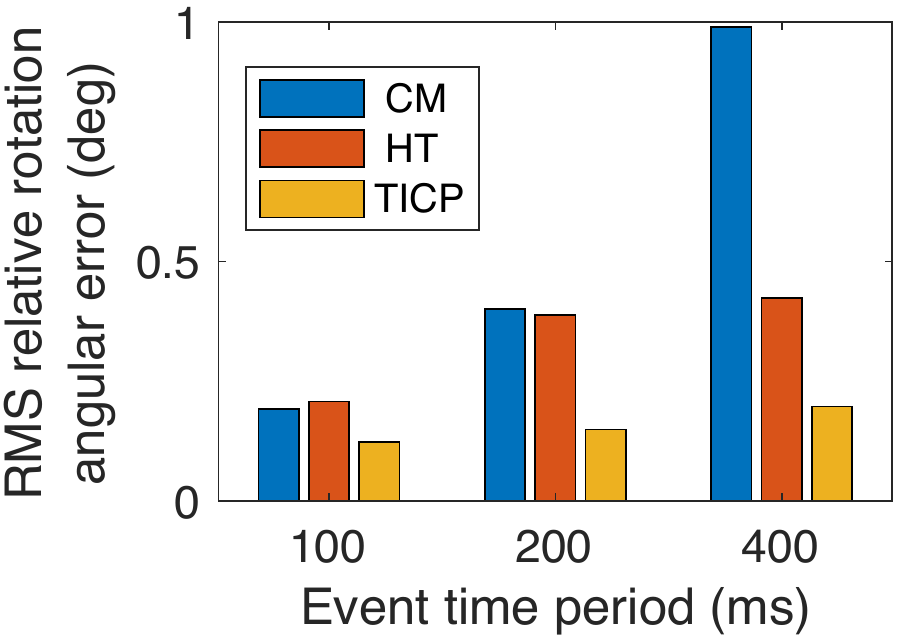}\label{fig:tokyoRR_}}\hspace{1em}
\subfigure[Sequence 6]{\includegraphics[width=0.31\textwidth, height=0.1775\textheight]{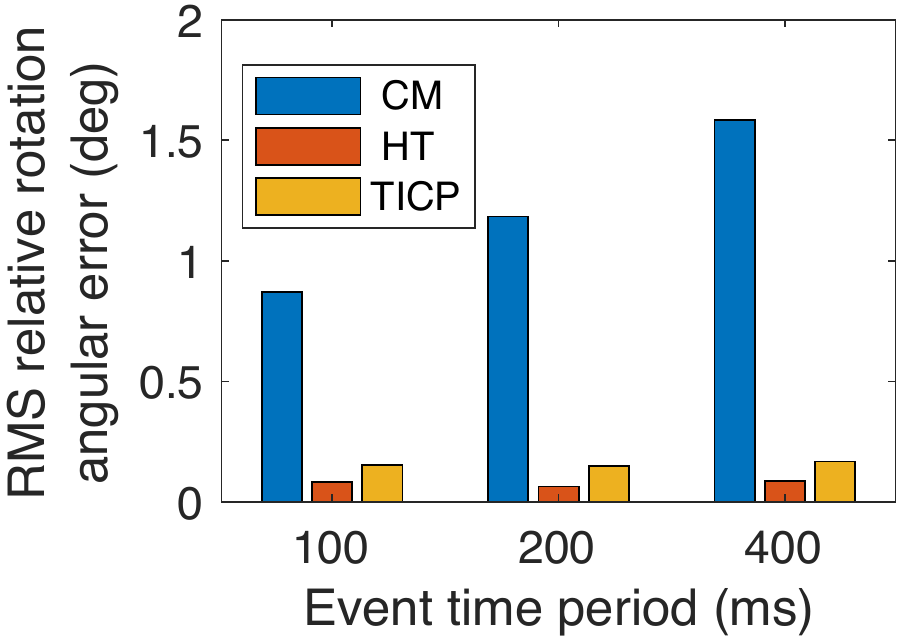}\label{fig:rioRR_}}
\caption{Comparison of relative rotation estimation accuracy using three methods (HT, TICP and CM) on 6 event streams/sequences.}
\label{fig:sixseqsRR}
\end{figure*}

Following~\cite{chin18}, we formulate rotation averaging as
\begin{align}\label{eq:augrotavg}
\begin{aligned}
& \underset{\{ \bR_t\}_{t \in \mathcal{T}}, \bR_{\mathcal{G}}}{\min}
& & \sum_{\langle \alpha,\beta \rangle \in \cN} \left\| \bR_\alpha -  \tilde{\bR}_{\alpha,\beta} \bR_\beta  \right\|_F\\
& & & + \alpha \sum_{\bR_\gamma \in \cM} \left\| \bR_\gamma - \tilde{\bR}_\gamma \bR_{\mathcal{G}} \right\|_F \\
& \text{subject to}
& & \bR_{\mathcal{G}} = \bI,
\end{aligned}
\end{align}
where $\bR_{\mathcal{G}}$ is a ``dummy" attitude variable, $\bI$ is the identity matrix, and $\alpha$ is a positive constant that defines the relative importance of the relative and absolute rotations. Intuitively, adding error terms of the form
\begin{align}
\left\| \bR_\gamma - \tilde{\bR}_\gamma \bR_{\mathcal{G}} \right\|_F = \left\| \bR_\gamma - \tilde{\bR}_\gamma \right\|_F, \;\;\;\; \gamma \in \cM
\end{align}
encourage consistency between some of the attitude estimates and the measured absolute rotations, which is then propagated to the rest of the sequence. 

To solve~\eqref{eq:augrotavg}, we temporarily ignore the constraint $\bR_{\mathcal{G}} = \bI$ optimise the attitudes using an existing rotation averaging algorithm (we used~\cite{chatterjee2013efficient} in our work). Then, right multiply each of the optimised absolute attitudes $\hat{\bR}_t$ with $\hat{\bR}_{\gamma}^{-1} = (\hat{\bR}_{\gamma})^{T}$ to re-orient the system. It has been shown that rotation averaging is quite insensitive to initialisations~\cite{olsson11,eriksson18}, thus, when solving~\eqref{eq:augrotavg} we simply initialise all rotation variables as the identity matrix.

In a practical system, instead of solving ever-growing instances of~\eqref{eq:augrotavg}, rotation averaging can be executed over a fixed-sized temporal sliding window.

\section{Results}\label{sec:results}

To evaluate the proposed event-based star tracking technique, we used the event data from~\cite{chin18,startrackdata}, where there are 11 event streams from observing star fields using an iniVation Davis 240C event camera (see details in~\cite{chin18} on data generation and event camera calibration). Each event stream contains ground truth absolute attitudes across the full stream.

Our evaluation focussed on the accuracy and runtime of relative attitude estimation and absolute attitude estimation. For both experiments, we compared against:
\begin{itemize}[leftmargin=1em,topsep=1pt,parsep=0em,itemsep=0em]
\item The baseline method of~\cite{chin18} that conducts robust registration using TICP on event images to estimate relative rotations (henceforth ``TICP").
\item The state-of-the-art event-based motion estimation algorithm of contrast maximisation~\cite{gallego18} (henceforth, ``CM").
\end{itemize}
See Sec.~\ref{sec:prevevents} for a summary of these methods. Due to space contraints, we were able to fit the results of only 6 event streams; see supplementary material for more results.

To compare estimated $\tilde{\bR}$ and ground truth $\bR^\ast$ rotations in our evaluation, we used the angular distance~\cite{hartley13}
\begin{align}
\angle(\tilde{\bR},\bR^\ast) = 2\arcsin\left(\frac{2}{\sqrt{2}} \| \tilde{\bR} - \bR^\ast \|_F\right).
\end{align}

\subsection{Accuracy of relative attitude estimation}

For each event stream, as per the multi-resolution architecture in Fig.~\ref{fig:overall}, we estimated relative rotations from event ``chunks" using HT, TICP and CM. Since each stream is about $45s$ long, there are 224 chunks of $400ms$ long, 449 chunks of $200ms$ long, and 899 chunks of $100ms$ long.

To objectively test the efficacy of the algorithms, we executed them on a common platform: Matlab on an Intel i7 2.7 GHz machine, and using \texttt{fmincon} to solve CM~\eqref{eq:contmax}. Since our focus here is relative rotation accuracy, it was not necessary to execute HT in a streaming fashion.

To assess the accuracy of a relative rotation estimate $\tilde{\bR}_{\alpha,\beta}$ from a time period $\bbT = [\alpha,\beta]$, we computed the angular distance of $\tilde{\bR}_{\alpha,\beta}$ to the ground truth relative rotation
\begin{align}
\bR^\ast_{\alpha,\beta} = \bR_{\alpha}^\ast (\bR_{\beta}^\ast)^T,
\end{align}
where $\bR_{\alpha}^\ast$ and $\bR_{\beta}^\ast$ ground truth absolute attitudes. Fig.~\ref{fig:sixseqsRR} shows the root mean square (RMS) angular distance
\begin{align}
\sqrt{ \sum_{\langle \alpha,\beta \rangle} \angle( \tilde{\bR}_{\alpha,\beta},\bR^\ast_{\alpha,\beta}   )^2 }
\end{align}
separated according to the length of the time periods $\bbT$ (200, 300 and 400 ms). The RMS error of HT is conclusively lower than CM on most of the input cases, followed by TICP which has lower error than HT in 4 of the 6 sequences.

\begin{figure*}[t]\centering
\subfigure[Sequence 1]{\includegraphics[width=0.32\textwidth, height=0.188\textheight]{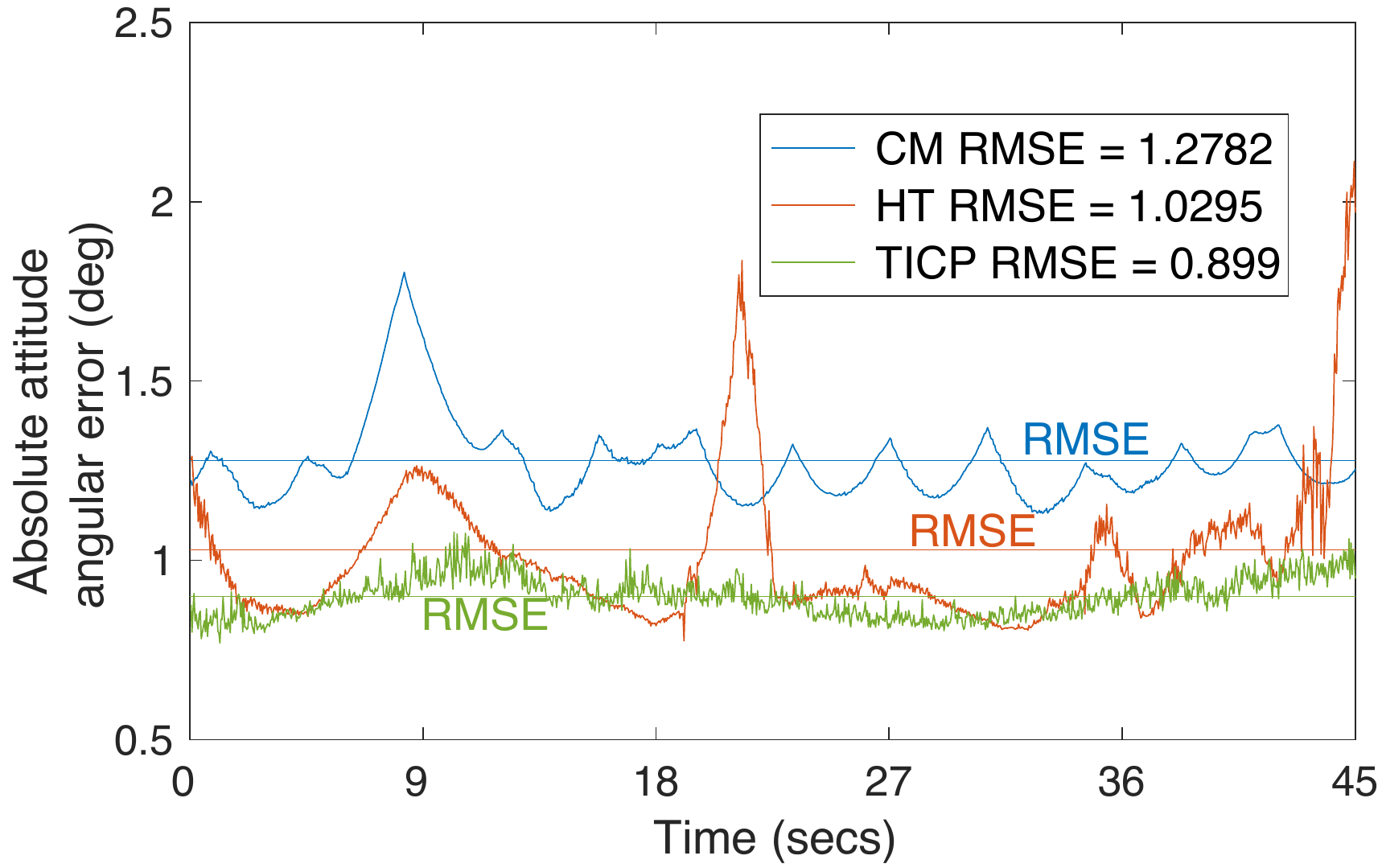}\label{fig:adelaideAR_}}\hspace{0.5em}
\subfigure[Sequence 2]{\includegraphics[width=0.32\textwidth, height=0.188\textheight]{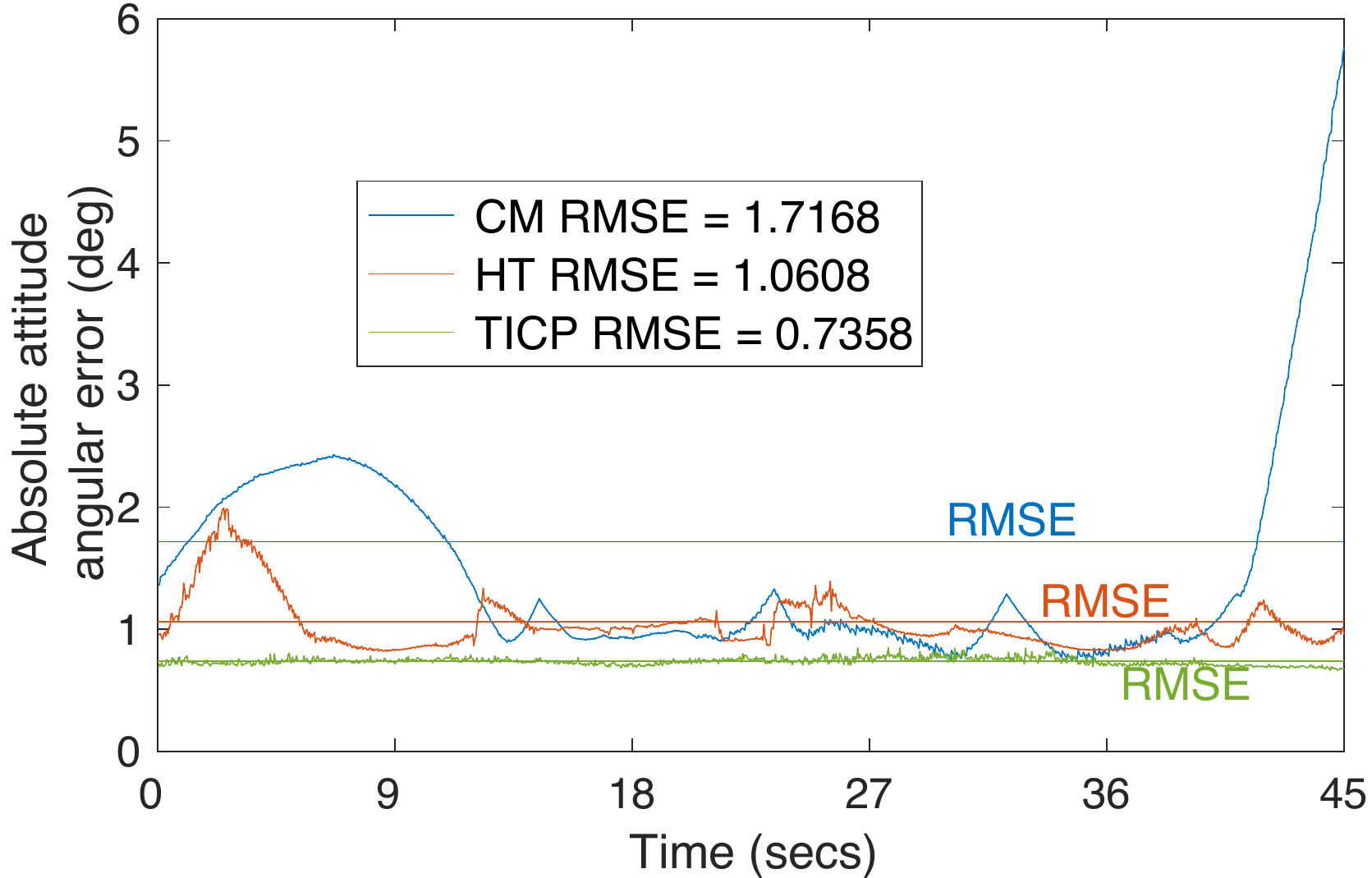}\label{fig:capetownAR_}}\hspace{0.5em}
\subfigure[Sequence 3]{\includegraphics[width=0.32\textwidth, height=0.188\textheight]{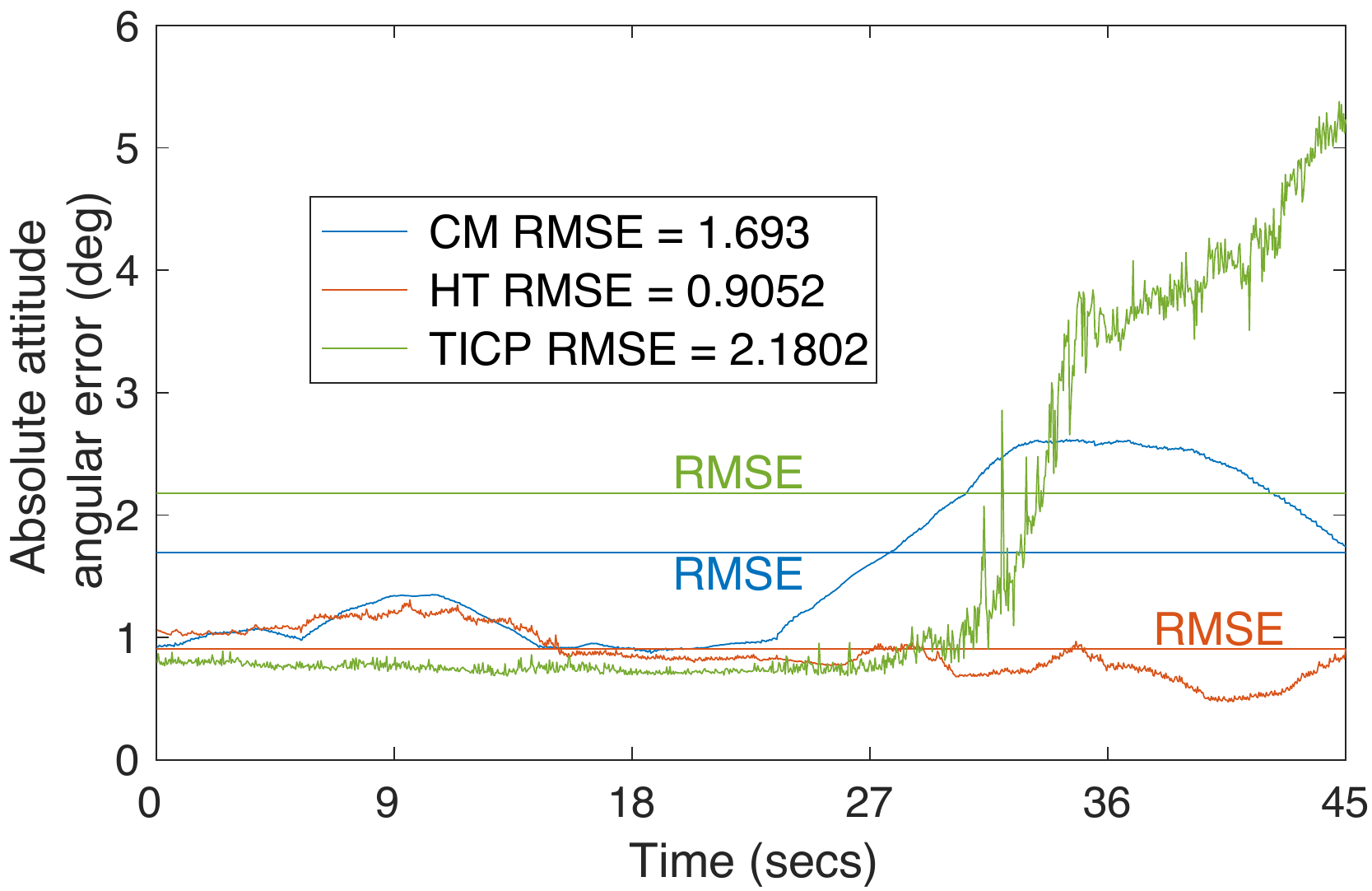}\label{fig:ecuadorAR_}}
\subfigure[Sequence 4]{\includegraphics[width=0.32\textwidth, height=0.188\textheight]{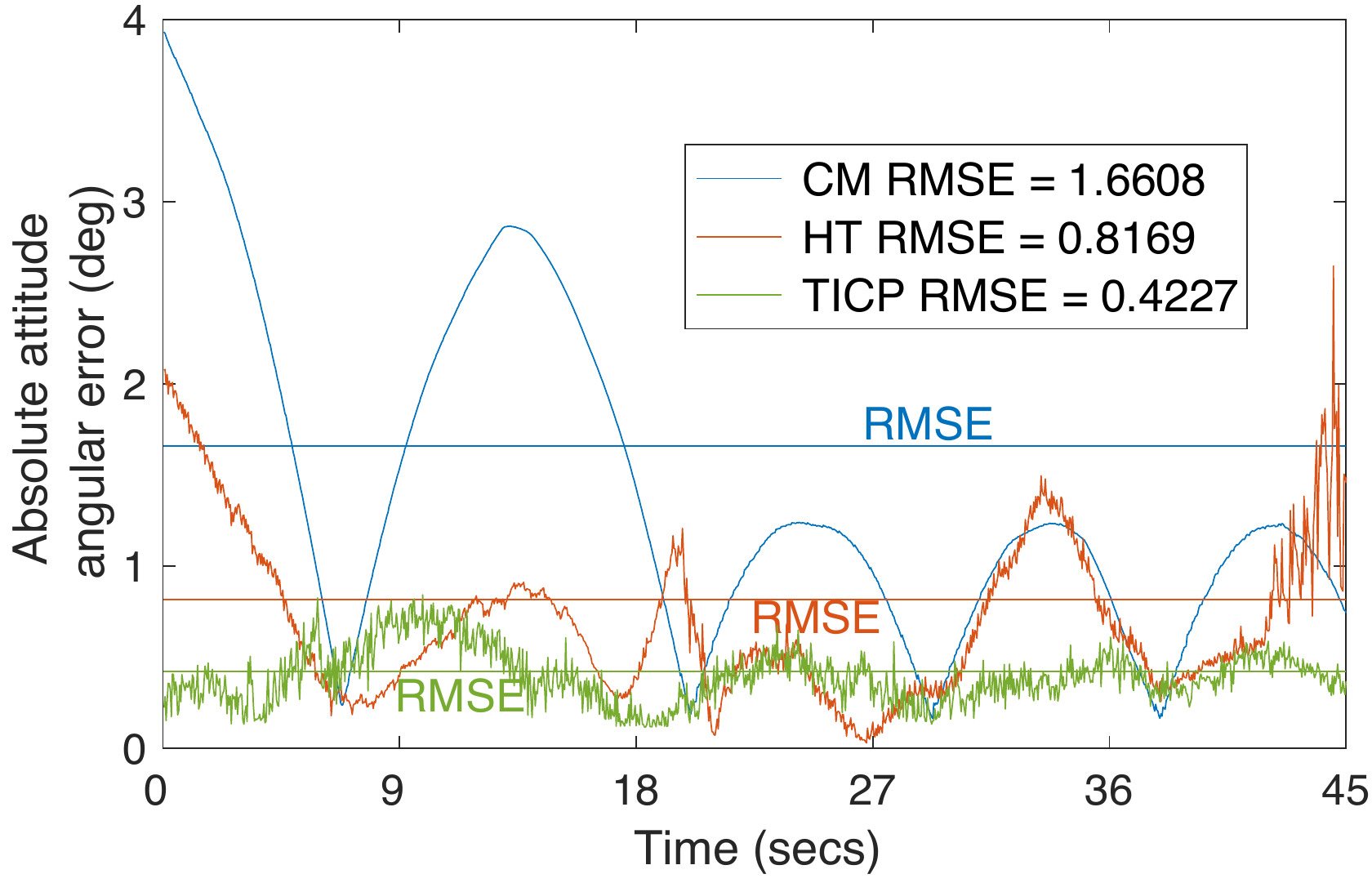}\label{fig:aucklandAR_}}\hspace{0.5em}
\subfigure[Sequence 5]{\includegraphics[width=0.32\textwidth, height=0.188\textheight]{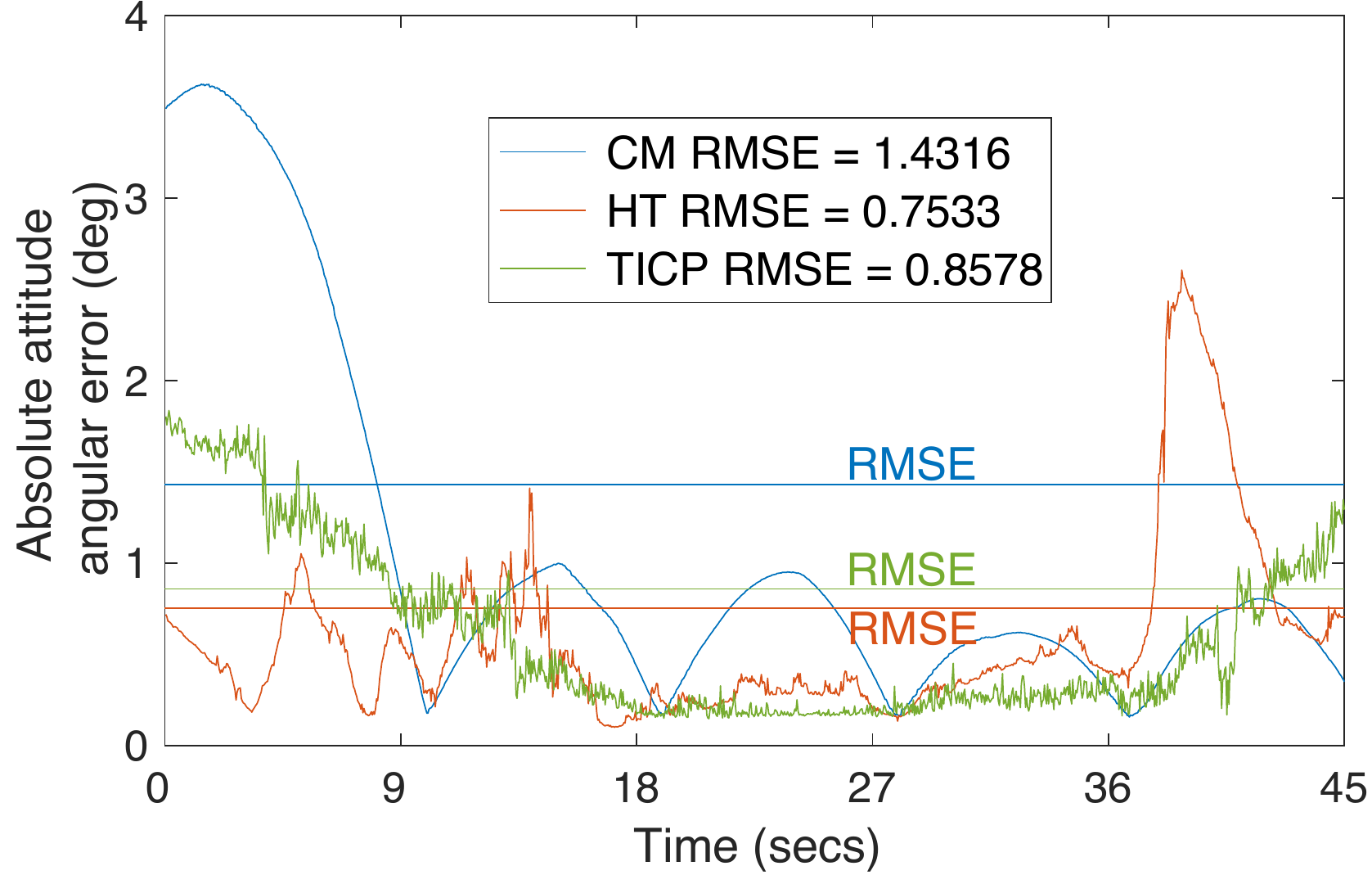}\label{fig:tokyoAR_}}\hspace{0.5em}
\subfigure[Sequence 6]{\includegraphics[width=0.32\textwidth, height=0.188\textheight]{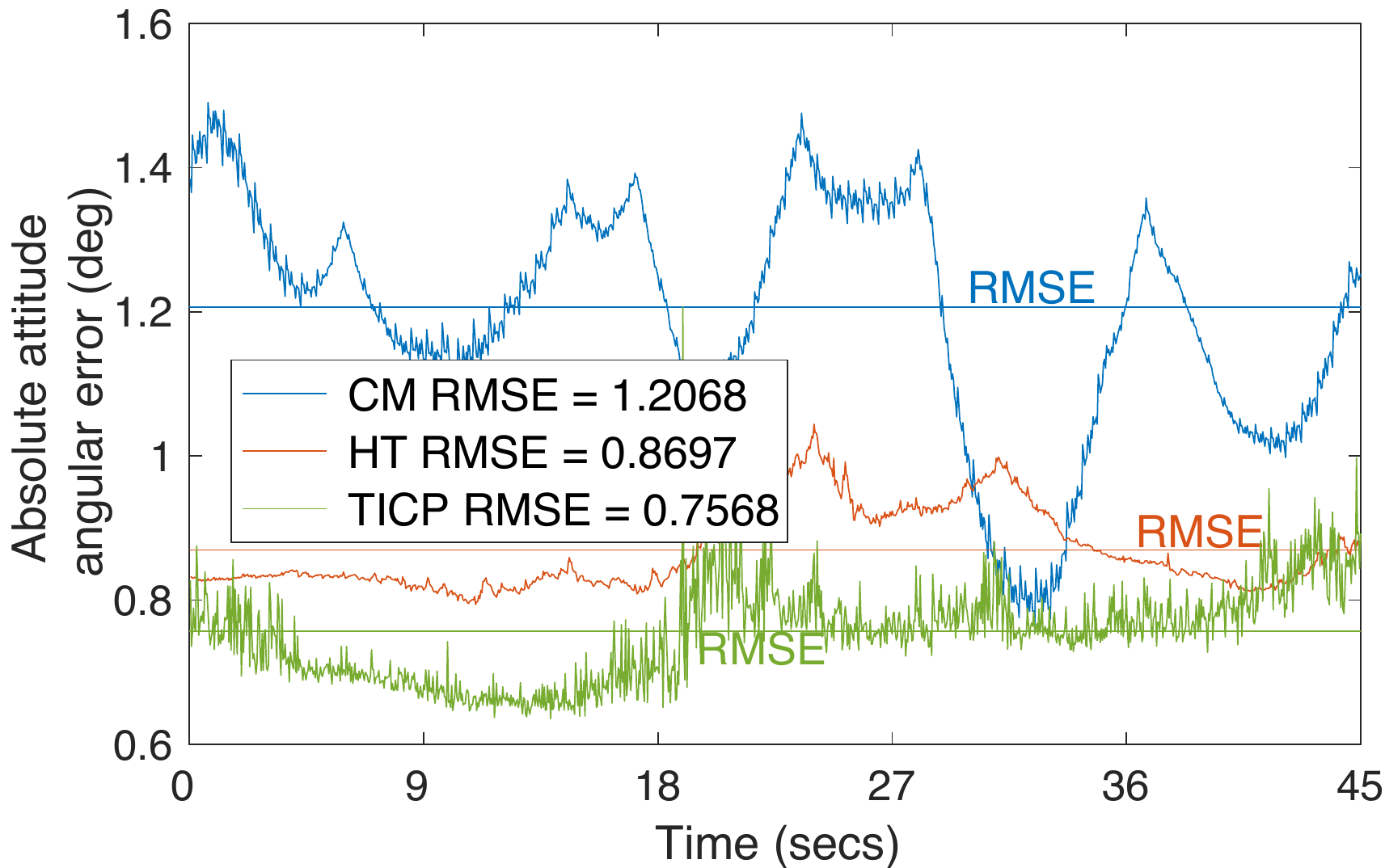}\label{fig:rioAR_}}
\caption{Comparison of absolute attitude estimation accuracy using three methods (HT, TICP and CM) on 6 event streams/sequences.}
\label{fig:sixseqsAR}
\end{figure*}

It should be reminded, however, that TICP does not intrinsically support asynchronous processing of event streams; moreover, it depends on much more complex routines (sorting, nearest neighbour search, alternating optimisation, etc.). In contrast, HT is designed to be simple enough to be parallelised on standard hardware (e.g., FPGA and SoC). Thus, it is not surprising that TICP can be more accurate than HT. In any case, as we show later, the accuracy of the final attitude estimates of HT is very competitive.

\begingroup
\setlength{\tabcolsep}{2.80pt} 
\begin{table}[t]\centering
\begin{tabular}{llllllll}
\hline
$|\bbT|$ & Me- & \multicolumn{6}{c}{Sequence number} \\
(ms) & thod & 1 & 2 & 3 & 4 & 5 & 6 \\
\hline
\multirow{2}{*}{100} & HT & 1.23 & 1.85 & 1.49 & 2.11 & 1.31 & 1.67 \\
 & CM & 16.46 & 28.31 & 21.89 & 33.15 & 19.43 & 24.36 \\
\hline
\multirow{2}{*}{200} & HT & 1.88 & 2.39 & 2.01 & 2.14 & 1.93 & 2.09 \\
 & CM & 45.68 & 61.45 & 53.46 & 65.78 & 48.91 & 61.24 \\
\hline 
\multirow{2}{*}{400} & HT & 2.51 & 2.98 & 2.67 & 3.11 & 2.63 & 2.83 \\
 & CM & 82.45 & 114.78 & 103.56 & 118.32 & 89.12 & 109.25 \\
\hline
\end{tabular}
\caption{Runtime (in seconds) of event-based processing methods (HT, CM) for relative rotation estimation on six event streams.}
\label{fig:runtimes}
\end{table}
\endgroup

Table~\ref{fig:runtimes} presents the average runtime taken by the event-processing methods to compute relative rotations. Note that as the time period increases, the runtime of CM increases much more rapidly than HT. While a better solver than \texttt{fmincon} can potentially speed up CM, even with an order of magnitude speedup, HT is still faster than CM. Note that the runtime of TICP does not change with the size of the event chunk, since it always aligns two event images of the same dimensions; due to this different computational paradigm, the runtime of TICP is not shown in Table~\ref{fig:runtimes}.

\subsection{Accuracy of absolute attitude estimation}

Fig.~\ref{fig:sixseqsAR} plots the ``raw" and RMS angular distances between ground truth and estimated absolute attitudes, based on performing the proposed rotation averaging technique (Sec.~\ref{sec:overall}) on the relative rotations (computed by HT, TICP and CM) and a small set of ``grounding" absolute attitudes~\eqref{eq:grounding} (5 per event stream, distributed uniformly across 45 seconds, and computed using~\cite{lang10,astrometry} on the motion compensated event image by HT on 100 ms chunks). As can be seen, the optimised absolute attitudes using HT and TICP are competitive enough with commercial star trackers~\cite{opromolla2017new} ($\ 1^\circ$ RMS angular error). The cost of rotation averaging is also low, i.e., 2 to 3 seconds per event stream of 45 seconds.

Towards the end of Sequence 3, the error of TICP seems to increase drastically. This was because there were much fewer stars in the FOV in that period, which reduced the relative rotation accuracy of TICP. This unfavourable situation did not affect the proposed HT method.

\section{Conclusion}\label{sec:conclusion}

In this paper, we proposed an event-based star tracking algorithm. The main aspect in our pipeline is a novel multiresolution asynchronous HT technique to accurately and efficiently estimate relative rotations on event streams. Our results show that our technique is superior to existing event-based processing schemes for motion estimation.

\clearpage


{\small
\bibliographystyle{ieee}
\bibliography{startracking}
}

\end{document}